\documentclass{article}

\usepackage{arxiv}

\usepackage[utf8]{inputenc} 
\usepackage[T1]{fontenc}    
\usepackage{hyperref}       
\usepackage{url}            
\usepackage{booktabs}       
\usepackage{amsfonts}       
\usepackage{nicefrac}       
\usepackage{microtype}      
\usepackage{lipsum}

\usepackage[nolist]{acronym}

\usepackage{paralist}

\usepackage{etoolbox}

\usepackage{url}
\makeatletter
\g@addto@macro{\UrlBreaks}{\UrlOrds}
\makeatother

\usepackage{multirow,tabularx}
\usepackage[export]{adjustbox}
\usepackage{calc}

\usepackage{amsmath}
\usepackage{amsfonts}
\usepackage{amssymb}

\usepackage{cleveref}

\usepackage{xspace}

\usepackage{ifthen}

\usepackage{subfigure}
\newboolean{showNotes}
\setboolean{showNotes}{true}

\DeclareRobustCommand{\notesAtabak}[1]{\ifthenelse {\boolean{showNotes}}
 {\textbf{\textcolor{blue}{Atabak: #1}}}
 {}
}
\DeclareRobustCommand{\notesManuel}[1]{\ifthenelse {\boolean{showNotes}}
 {\textbf{\textcolor{ForestGreen}{Manuel: #1}}}
 {}
}
\DeclareRobustCommand{\notesPlinio}[1]{\ifthenelse {\boolean{showNotes}}
 {\textbf{\textcolor{magenta}{Plinio: #1}}}
 {}
}
\DeclareRobustCommand{\General}[1]{\ifthenelse {\boolean{showNotes}}
 {\textbf{\textcolor{BurntOrange}{General: #1}}}
 {}
}

\DeclareRobustCommand{\hhref}[1]{\texttt{\url{#1}}}

\makeatletter
\DeclareRobustCommand\onedot{\futurelet\@let@token\@onedot}
\def\@onedot{\ifx\@let@token.\else.\null\fi\xspace}

 \def\Eg{\emph{E.g}\onedot}
\def\ie{\emph{i.e}\onedot}

\def\etal{\emph{et al}\onedot}
\makeatother

\begin{acronym}
\acro{COCO}{Common Objects in COntext}
\acro{CNN}{Convolutional Neural Network}
\acro{PCA}{Principal Component Analysis}
\acro{mAP}{Mean Average Precision}
\acro{MAE}{Mean Absolute Error}
\acro{VAE}{Variational Auto Encoder}
\acro{MLP}{Multi Layer Perceptron}
\acro{GAN}{Generative Adversarial Network}
\acro{LSTM}{Long Short Term Memory}
\acro{RGB-D}{RGB-Depth}
\acro{KL}{Kullback-Leibler}
\acro{MDN}{Mixture Density Network}
\acro{RNN}{Recurrent Neural Networks}
\acro{PSNR}{Peak Signal to Noise Ratio}
\acro{SSIM}{Structural Similarity}
\acro{FVD}{Fr{\'e}chet Video Distance}
\acro{NCCL}{Normalized Cross Correlation Loss}
\acro{PCDL}{Pairwise Contrastive Divergence Loss}
\acro{CDNA}{Convolutional Dynamic Neural Advection}
\acro{SAVP}{Stochastic Adversarial Video Prediction}
\acro{SVG-LP}{Stochastic Video Generation with Learned Prior}
\acro{ConvLSTM}{Convolutional Long Short Term Memory}
\acro{BAIR}{Berkeley Artificial Intelligence Research Lab}
\acro{LPIPS}{Learned Perceptual Image Patch Similarity}
\acro{VP}{Video Prediction}
\acro{QoE}{Quality of Experience}
\acro{MOS}{Mean Opinion Score}
\acro{SV2P}{Stochastic Variational Video Prediction}

\end{acronym}

\title{Action-conditioned Benchmarking of Robotic Video Prediction Models: a Comparative Study}

\author{
  Manuel Serra Nunes$^{1}$ \qquad
  Atabak Dehban$^{1,2}$ \qquad
  Plinio Moreno$^{1}$ \qquad
  Jos\'e Santos-Victor$^{1}$\\
  $^{1}$Institute for Systems and Robotics, Instituto Superior T\'ecnico, Lisbon, Portugal\\
  $^{2}$Champalimaud Centre for the Unknown, Lisbon, Portugal\\
  \texttt{\{mserranunes,adehban,plinio,jasv\}@isr.tecnico.ulisboa.pt} \\
}

\begin{document}
\maketitle
\begin{adjustwidth}{-0.703cm}{-0.527cm}
\begin{multicols}{2}
\begin{abstract}
A defining characteristic of intelligent systems is the ability to make action decisions based on the anticipated outcomes.
Video prediction systems have been demonstrated as a solution for predicting how the future will unfold visually, and thus, many models have been proposed that are capable of predicting future frames based on a history of observed frames~(and sometimes robot actions).
However, a comprehensive method for determining the fitness of different video prediction models at guiding the selection of actions is yet to be developed. 

Current metrics assess video prediction models based on human perception of frame quality.
In contrast, we argue that if these systems are to be used to guide action, necessarily, the actions the robot performs should be encoded in the predicted frames.
In this paper, we are proposing a new metric to compare different video prediction models based on this argument.
More specifically, we propose an action inference system and quantitatively rank different models based on how well we can infer the robot actions from the predicted frames.
Our extensive experiments show that models with high perceptual scores can perform poorly in the proposed action inference tests and thus, may not be suitable options to be used in robot planning systems.
\end{abstract}

{ \centering
  \includegraphics[scale=0.26]{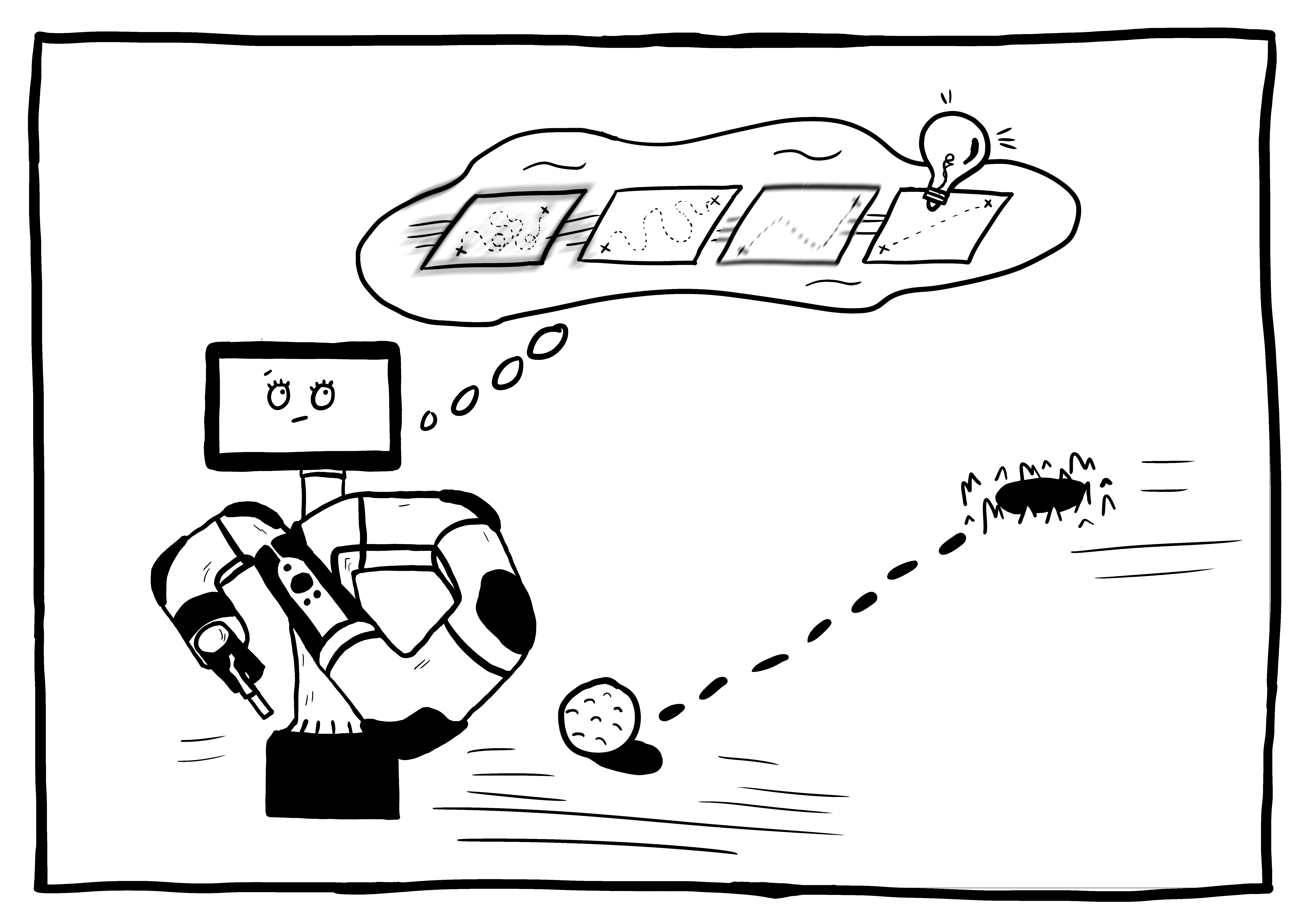}
  \captionsetup{width=.8\linewidth}
  \captionof{figure}{
  A sawyer robot imagining different possible outcomes of executing an action in the hypothetical task of pushing a ball to a whole. Image credit: Teresa Serra Nunes.}
  \label{fig:sawyer}
}

\end{multicols}
\end{adjustwidth}

\section{Introduction}
\label{sec:intro}

An important stepping stone on the path towards intelligent robotic agents is providing them with the ability to explore their environment and to learn from interaction.
Visual data, in the form of video, plays a central role in this problem and has led to great success in problems such as unsupervised learning of object keypoints~\cite{kulkarni2019unsupervised} and action recognition \cite{simonyan2014two}.
In this direction, the next step should be for the robot to be able to learn the inherent workings of a real world environment and to understand how different bodies move, deform and influence each other.

As suggested by~\cite{srivastava2015unsupervised}, a notion that might lead to cracking this challenge is that, if based on an observed sequence of visual queues the robot has the ability to predict the imminent future, then it must have acquired a representation of the spatial and temporal dynamics of the world.
Predicting future video frames is perhaps the most straightforward materialization of this idea as the better a system can make predictions about future observations, the better the acquired feature representation must be~\cite{mathieu2015deep}.
For example, a robot that is able to predict that a falling stick will become occluded by a box, must understand where the trajectory of the stick will take it, be capable of perceiving depth and also of recognizing that the box in the foreground is opaque.

From a robotic perspective, if the predictions consider the actions of the agent itself, \ie are conditioned on the action, then the representation should also help understand how performing an action in a situation will affect the future appearance of the scene and thus, guide action decisions, an idea illustrated in \cref{fig:sawyer}.

Similarly, the idea of anticipating sensory inputs to optimize action response in the human brain is studied under the theory of prospective coding, which explains the phenomenon by which representations of future states influence event perception and generation~\cite{friston2005theory,schutz2007prospective}.
Forward generative models constitute a fundamental part of predictive coding theory, especially in the domain of human action planning~\cite{fitzgerald2015active} where the latency between the stimulus of the retina and the corresponding acknowledgement by the responsible region of the brain makes it difficult to select appropriate actions in response to rapidly evolving events.

The capabilities of \ac{VP} systems to serve as forward models have been explored in~\cite{ha2018world}, with the introduction of an architecture that learns a policy for solving OpenAI Gym~\cite{brockman2016openai} reinforcement learning problems using an encoder of observed video frames and a MDN-RNN to predict future visual codes, given current and past observations and executed actions.
In the context of robotic planning in a real world scenario \cite{finn2017deep} uses a \ac{VP} model to continuously sample the expected future given different sequences of actions.
The sequence that maximizes the likelihood of the robot achieving the goal of pushing an object to a specified location is selected at each time step to be executed.

Naturally, in these applications, the ability to select the best possible action is very dependent on how well the \ac{VP} model can anticipate future observations based on the robot actions and the current status of the scene.
Having a metric that can rank video prediction models based on how well they perform as a forward model is therefore of fundamental importance.

As will be described in \cref{sec:related}, most state-of-the-art work in video prediction measures the performance of the models using metrics designed to reflect human perception of quality.
While these metrics might be useful for applications such as precipitation nowcasting \cite{xingjian2015convolutional} or semantic segmentation prediction~\cite{luc2017predicting}, we argue that they are not necessarily adequate in action oriented applications such as robotic planning where the quality of the video prediction model should be measured by how well it can guide action decisions from the predicted frames.

Inspired by this notion, we propose a new, simple metric for ranking video prediction models from a robotic standpoint.
Given a sequence of predicted frames, we train a model to infer the action performed at each time step.
The ability of our inference machine to recognize the correct sequence of actions only from the predicted frames should indicate that the representation of the world held by the video prediction model is correctly encoding action features and it is able to understand the consequences of executing a given action at the current state of the environment.

This paper has the following three contributions:
\begin{itemize}
	\item we propose a novel action-based quality assessment metric for robotic video prediction models;
	\item we apply the metric on several different models and quantitatively rank them from a robotic action-inference perspective;
	\item we qualitatively compare our method with other metrics and show that in most cases our quality measure can independently assess models, providing new insights about their performance as action-conditioned video prediction models.
\end{itemize}

In addition, we provide the implementation of our experiments to facilitate assessment of \ac{VP} models that we did not consider or that might be proposed in the future\footnote{\url{https://github.com/m-serra/action-inference-for-video-prediction-benchmarking}}.

The rest of the paper is organized as follows: in \cref{sec:related} we examine the related work on different \ac{VP} models with an emphasis on the research directions that we consider in this paper.
In \cref{sec:methods} we explain the details of of our method, metrics we used, and several design choices we considered that made this work possible.
\Cref{sec:experiment} begins with a description of the dataset used in our experiments. It then continues by discussing how well different methods could compete in our metric and how do they compare in terms of other metrics already developed in the literature.
Finally, we draw our conclusions and discuss promising future research directions in \cref{sec:conclusions}.

\section{Related Lines of Research}
\label{sec:related}

\subsection{Video Prediction}
The importance of anticipation and predictive sensori processing has long been regarded as crucial in controlling neural and cognitive processes such as perception, decision making and motion in both humans and animals, with studies on the subject dating as far back as the $19^{th}$ century~\cite{james_w} and extending into the $21^{st}$ century~\cite{pezzulo2014internally,rao1999predictive}.
In the field of robotics, these concepts inspired the development of sensori-motor networks~\cite{wolpert2011principles} which emulate the interaction between the visual and motor systems in organisms to predict future visual stimulus.
In \cite{santos2014sensori} sensori-motor networks are applied to small image patches to predict the next time step's stimulus.

However, when the problem is extended to more generic settings involving observations of a complete scene and longer temporal sequences, the high dimensionality of the data and the rich, complex, and uncertain dynamics of the real world become a bigger hurdle.
In recent years, research in neural networks has mitigated these problems, with the development of \acp{CNN}, that reduce the dimensionality burden in image processing, \acp{RNN} which capture the information contained in sequential data and a combination of the two in the \acp{ConvLSTM}~\cite{xingjian2015convolutional}.
All these systems have been widely adapted in the field of video prediction.

Influential work on video prediction by Mathieu~\etal~\cite{mathieu2015deep} focused on improving the quality of the generated video by experimenting with different loss functions as an alternative to $\ell_2$ which is hypothesized to cause blurry predictions.
One of the most meaningful contributions to video prediction was perhaps the introduction of the concept of pixel motion by Finn~\etal~\cite{finn2017deep} Xue~\etal~\cite{xue2016visual} and De Brabandere~\etal~\cite{jia2016dynamic} which liberates the system from having to predict every pixel from scratch by modelling pixel motion from previous images and applying it to the most recent observation.
Since then several authors have continued work in this direction:~\cite{babaeizadeh2017stochastic} tries to account for the stochasticity of the world by conditioning the predictions on stochastic variables while~\cite{lee2018stochastic} explores how the introduction of a \ac{GAN} improves the visual quality of predictions.

Other lines of research have included motion and content decomposition~\cite{villegas2017decomposing}, predicting transformations on feature maps~\cite{oh2015action,xue2016visual}, and biologically inspired approaches like Lotter~\etal~\cite{lotter2016deep} which proposes a hierarchical architecture that emulates the top-down and bottom-up transmission of local predictions and errors in predictive coding theories of human perception.

In this work we focus on action-conditioned video prediction models as it is presumable that those are the most suited models for use in robotic planning.
We select
\begin{inparaenum}
  \item \ac{CDNA}: a deterministic model based on pixel-motion modelling~\cite{finn2016unsupervised},
	\item \ac{SAVP}: which also models pixel motion but introduces variational and adversarial terms to the loss, to try to improve prediction quality and account for the variability in the environment~\cite{lee2018stochastic},
	\item	a variant of \ac{SAVP} in which the adversarial term is suppressed,
	\item \ac{SV2P} \cite{babaeizadeh2017stochastic}, which builds on top of \ac{CDNA} to account for the world's stochasticity 
	\item and finally we test \ac{SVG-LP}: the stochastic, action-free model of \cite{denton2018stochastic}.
\end{inparaenum}

\subsection{Assessment of video prediction models}

A common trend across the described work on video prediction is the evaluation of model performance based on metrics designed to mirror human perception of quality in image and video or, \ac{QoE}.
This is a subjective concept, which depends not only on the data fidelity of the reconstructed image or video but also on the personal experience and expectations of the viewer \cite{winkler2008evolution}.
The standard measure for \ac{QoE} is the \ac{MOS} which is the average quality rating, given by a sample of viewers.
\ac{QoE} prediction is an active area of research in which proposed methods are usually compared to the \ac{PSNR} benchmark.
\ac{PSNR} is a logarithmic measure of the mean squared error between a distorted image and its original and its mathematical simplicity and desirable optimization properties make it the most popular metric for image quality.
However, \ac{PSNR} compares images pixel by pixel, not taking into account the content, leading to pathological cases~\cite{winkler2008evolution} in which it sometimes fails at approximating human judgement.

An alternative metric that addresses this problem is the \ac{SSIM} Index~\cite{wang2004video}, which starts from the principle that signals that are close in space have strong dependencies between each other and that the human visual system is highly adapted for extracting this structural information.
\ac{SSIM} indices are calculated using a sliding window of size $8\times 8$ which produces an index map.
The index is 1 if the structures of the two images within each patch are the same and the final \ac{SSIM} score corresponds to the average of the index map.
More recently, \ac{LPIPS} metrics, based on learned features of neural networks such as VGG have shown remarkable capabilities as a perceptual distance metric \cite{zhang2018unreasonable}.

Inspired by the developments in image generation, methods that are specifically designed for assessing realism in generated video have also been proposed~\cite{bhattacharjee2017temporal}.
\Eg \ac{FVD}~\cite{unterthiner2018towards} accounts for visual quality, temporal coherence, and diversity by measuring the distance between the distribution that originated the observed data and the distribution from which the predicted video is generated, instead of comparing pixels or image patches.

In this work we compare the performance of \ac{VP} models on our proposed metric with performance on \ac{PSNR} and \ac{SSIM}, which are the most commonly used metrics, and on \ac{FVD}.
\section{Methods}
\label{sec:methods}

In this section we present a simple method for ranking \ac{VP} models based on their capacity to guide a robotic agent's action decisions, reflected by the performance of action inference systems.
We start by assuming that the better the dynamics representation of the agent is at encoding action features, the better it will be for planning actions based on the expected outcome.
Under this assumption, the problem turns into evaluating how well a \ac{VP} model is encoding action features and assigning it a score based on such evaluation.

With this in mind, we hypothesise that the capacity to observe a sequence of predicted frames and infer the executed actions should be an indicator that the \ac{VP} model is correctly encoding action features.
To better illustrate the idea we may first consider a failure case: if the \ac{VP} model produces a future that could never stem from the context frames and actions, then the action inference model will likely recognize a set of actions in the predicted frames that do not correspond to the ground truth action sequence, resulting in a low action inference score.
On the other hand, if the \ac{VP} model understands the consequences of the input actions, then the frames it predicts should be in conformance with what ends up happening in practice, allowing the inference model to recognize the actual executed actions and achieve a high score.

\subsection{Video prediction}

In order to compare how the proposed metric correlates with \ac{PSNR}, \ac{SSIM} and \ac{FVD}, we start by selecting a group of \ac{VP} models from prior work to be tested using our metric.
By comparing model performance under metrics designed to predict human quality perception with our metric, designed to assess the capabilities of the model to guide action decisions, we intend to answer the question \textit{``Does a good video prediction from a human perspective correspond to a good video prediction from the standpoint of a robot making decisions?''}.
This is an interesting question considering that a change in model ranking, when compared to \ac{PSNR} or \ac{SSIM}, may not only influence the choice of the \ac{VP} in an action planning experiment but also indicate that the best representation for a robot to make a decision may not look like anything a human may recognize~\cite{inception,olah2018the} and inspire new lines of research such as optimizing for losses other than ground truth similarity.

The selection of the tested \ac{VP} models was made with the goal of covering the main approaches to video prediction, which opens up the possibility of identifying the most significant features of a \ac{VP} model used in a robotic planning context.
With that in mind, the selected models were \ac{CDNA}, \ac{SAVP}, SAVP-VAE, \ac{SV2P} and \ac{SVG-LP} which, with the exception of \ac{CDNA}\footnote{Since the pre-trained model was not available, we retrained the model according to the published article.}, were all tested on pre-trained versions provided by the authors.

\begin{figure}[h]
 \includegraphics[scale=0.48]{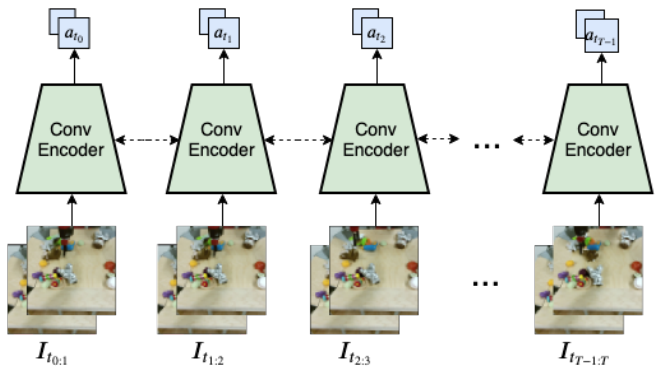}
 \centering
 \captionsetup{width=.78\linewidth}
 \caption{Action inference network unrolled in time. At each time step the network receives a pair of frames and outputs a multidimensional recognized action.}
 \label{fig:action_inference_network}
\end{figure}

\subsection{Action inference model}

To assess the quality of models, we first train a simple convolutional neural network to infer the actions executed between every two frames using predicted videos.
The actions are assumed to be continuous and multidimensional, to be representative of most robotic control action-spaces. Each pair of frames is concatenated along the channels dimension and given to the network as input, as illustrated in \cref{fig:action_inference_network}. Because action dynamics should not change over time, model parameters are shared across all time steps of a sequence. While an \ac{RNN} would typically have been useful for learning the sequence of executed actions, we choose to input two frames at a time, cutting off any temporal correlation between actions. This forces the inference model to identify actions from the frames instead of focusing  on learning the action distribution.
\section{Experiments and Results}
\label{sec:experiment}

\subsection{Experimental setup}
\begin{figure}[htbp]
\centering     
\subfigure[]{\label{fig:gripper_trajectory}\includegraphics[scale=0.26]{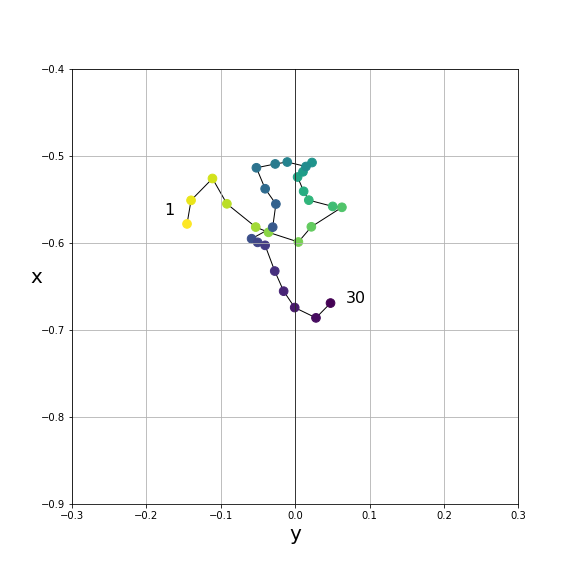}}
\qquad
\subfigure[]{\label{fig:data_distribution}\includegraphics[scale=0.26]{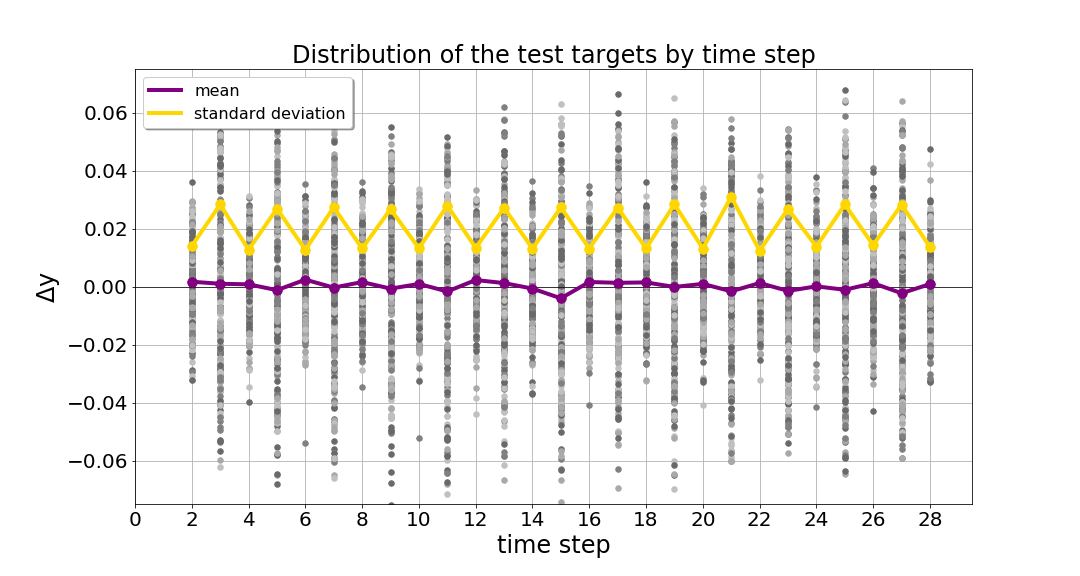}}
\captionsetup{width=.78\linewidth}
\caption{(a) A sample trajectory of the gripper illustrating the random nature of the actions. (b) Distribution of the test targets $\Delta y$, revealing a characteristic alternating standard deviation.}
\end{figure}

We conduct our experiments using the BAIR robot push dataset~\cite{ebert2017self} which consists of a robotic arm pushing a collection of objects inside a table.
This dataset was collected in the context of visual planning and video prediction and has since become a benchmark in the field~\cite{denton2018stochastic,castrejon2019improved,ebert2018visual,kumar2019videoflow}.
The dataset contains 43520 examples of random movements, as exemplified by a birds eye view of the gripper trajectory during a sample in \cref{fig:gripper_trajectory}.
Videos are 30 frame long, with $64 \times 64$ RGB images, collected at 10 Hz.
The dataset also provides the commanded actions sequences, a 4-dimensional array representing the joint velocities and whether the gripper is open or closed, and a 3-dimensional array representing the Cartesian coordinates of the gripper.

All tested \ac{VP} models were pre-trained by the respective authors with exception of \ac{CDNA}, which was trained on over $200000$ iterations, using scheduled sampling~\cite{bengio2015scheduled}.
At training time, models receive 2 context frames and actions (with the exception of \ac{SVG-LP} which only receives the frames) and predict video up to time step 12, with each prediction being fed back as input for the next time step.

After training, a forward pass is made over the entire training set and the generated predictions, this time generalizing until step 30, are saved as a new dataset for subsequent training of the action inference model.
Having a dataset of predictions for each \ac{VP} model, the action inference network is trained on the 28 frame long predictions.
In our experiments, we define the actions being inferred as the displacements $\Delta x$ and $\Delta y$ of the robot's gripper along the $x$ and $y$ axis, between every two time steps.
The ground truth targets for the actions are directly extracted from the BAIR dataset gripper state sequences by subtracting consecutive temporal positions on the $x$ and $y$ axis.
This results in an action sequence of length 27 for each 28 frame predicted video.
Finally, an oracle model is obtained by training an inference model (\cref{fig:action_inference_network}) directly on ground truth frames from BAIR, and not on frames predicted by a VP model. The performance of the oracle on the BAIR test set is used as a reference.

A characteristic of the BAIR dataset which has particular effect on the results is the fact that joint velocities are only updated every two frames. Even though the gripper position still changes at every time step, the variance of the change, \ie the variance of $\Delta x$ and $\Delta y$ is higher on time steps in which joint velocities are updated. This aspect of the data is depicted in \cref{fig:data_distribution} where the $\Delta y$ targets of the test set are scattered, revealing an alternating standard deviation.
In practice, this alternating nature results in the action inference network not experiencing all types of actions the same way, therefore becoming better fit to some situations than others. For this reason, results are presented separately for odd and even frames  

\subsection{Quantitative Comparison}
\label{sec:experiments_quantitative}

\begin{figure}[htbp]
 \includegraphics[scale=0.23]{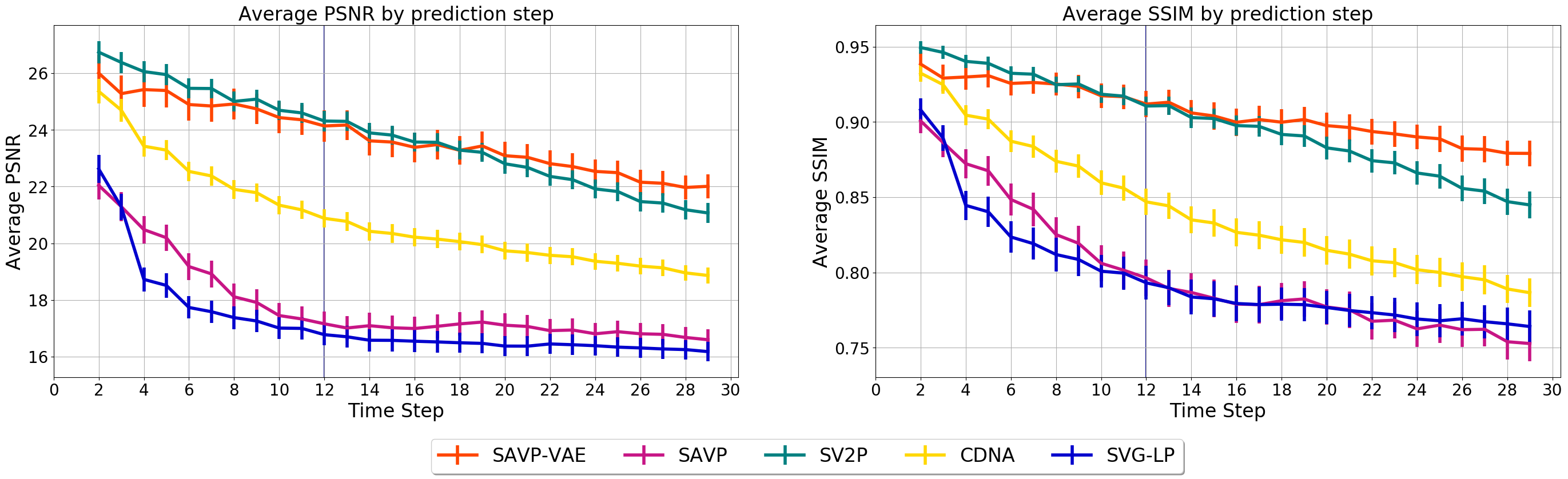}
 \centering
 \captionsetup{width=.78\linewidth}
 \caption{Average PSNR and SSIM over the test set with 95\% confidence interval. Results were reproduced with modification from \cite{finn2016unsupervised,denton2018stochastic,lee2018stochastic}.}
 \label{fig:cdna_ssim}
\end{figure}

We start by evaluating the selected group of \ac{VP} models on some of the traditionally used metrics described in section \ref{sec:related} and on the recently proposed \ac{FVD}.
As opposed to the methodology adopted by some of the previous work~\cite{lee2018stochastic,denton2018stochastic} where 100 possible futures are sampled and the best score of the group is reported, we choose to sample a single time, in order to better approximate the conditions of a robot planning actions.
This approach has especial impact on action-free models like \ac{SVG-LP}, that are exposed to greater uncertainty.
Regarding the action-conditioned models, the results displayed in \cref{fig:cdna_ssim} are in line with previous reports, indicating that models have better performance when conditioned on both actions and stochastic variables, as is the case with SAVP-VAE and \ac{SV2P}. On the other hand, the addition of an adversarial loss term seems to affect performance negatively, which reflects on \ac{SAVP} having a lower \ac{PSNR}/\ac{SSIM} score than a deterministic model like \ac{CDNA} despite the high visual appeal of the predicted frames.

We compute the \ac{FVD} scores for the test set predictions in \cref{table:r2_l1_fvd} using batches of size 32 and discard the two context frames to only consider the predictions of length 28.
This approach is different from the one proposed in~\cite{unterthiner2018towards}, therefore resulting in higher \ac{FVD} scores but preserving model rankings.

\begin{figure}[h]
 \includegraphics[scale=0.23]{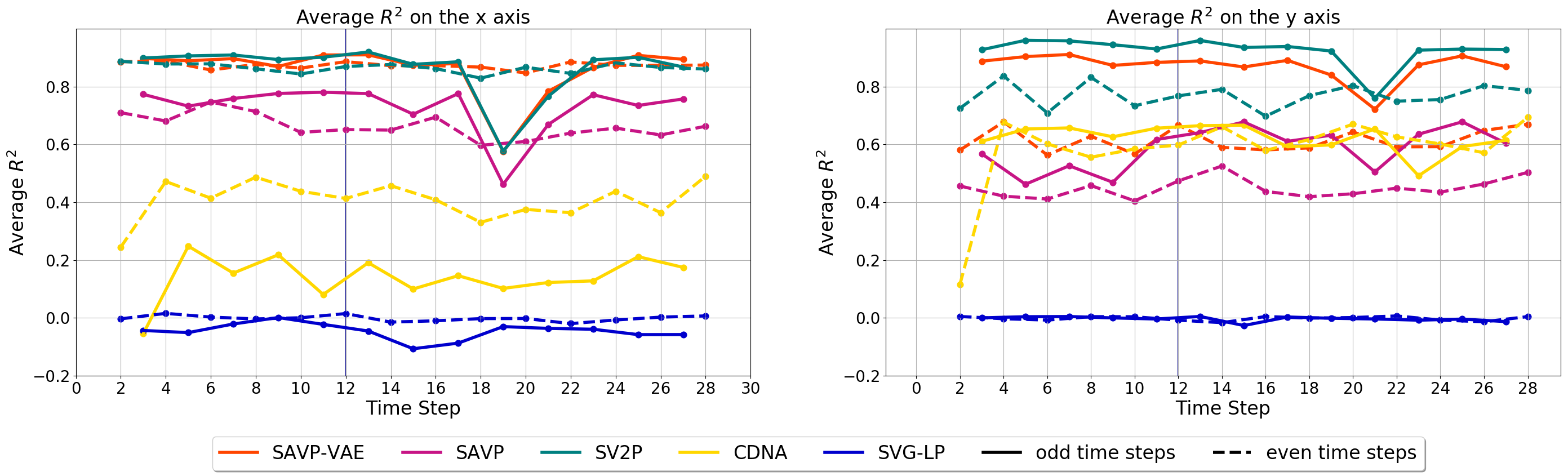}
 \centering
 \captionsetup{width=.78\linewidth}
 \caption{$R^2$ results over time for predictions made by different \ac{VP} models. Odd and even time steps are shown separately.}
 \label{fig:r2_plots}
\end{figure}

\begin{figure}[htbp]
 \includegraphics[scale=0.23]{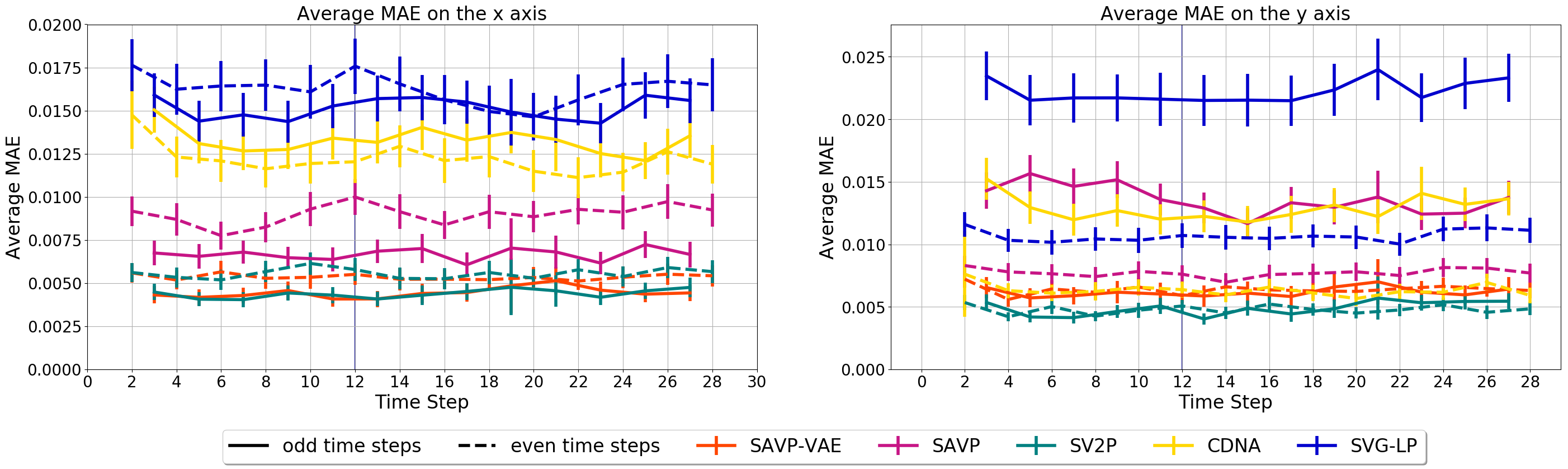}
 \centering
 \captionsetup{width=.78\linewidth}
 \caption{MAE results over time for predictions made by different \ac{VP} models. Odd and even time steps are shown separately.}
 \label{fig:l1_plots}
\end{figure}

For each \ac{VP} model's predictions dataset, the action inference model that produces the best validation score during training is selected. To measure how well it can identify the executed actions, we compute the $R^2$ goodness-of-fit metric which in our particular case represents the percentage of change in actions variances that is explained by the inferences.
A model that perfectly identifies the executed actions will have a score of $1.0$ whereas a model that simply outputs the mean $\Delta x$ and $\Delta y$ will have a score of $0.0$.
It is worth noting that while $R^2$ may not be a strong enough statistic for comparing regression models, the focus of this work is to assess \ac{VP} models, using the same inference model.
In our experiments $R^2$ is computed along the 256 examples of each time step and the evolution of the metric over time is reported in \cref{fig:r2_plots}.
The \ac{MAE} is also presented in \cref{fig:l1_plots}, computed in the same way as $R^2$, for each time step.
The most immediate characteristic in the temporal evolution of action inference that arises from an initial analysis of figures \ref{fig:r2_plots} and \ref{fig:l1_plots} is that the temporal downgrade artefact in performance observed in \ac{PSNR} and \ac{SSIM} does not manifest in the capacity of the model to recognize the actions, with exception of results for \ac{CDNA}.
This quality of the metric stems from the fact that model parameters are shared across all time steps as action dynamics do not change over time and \ac{VP} models should have a consistent encoding for all time steps.
For this reason, a \ac{VP} model that encodes actions in a consistent manner should allow the inference network to better learn how to recognize actions and will therefore display stable $R^2$ and \ac{MAE} scores across time, as is verified for \ac{SAVP}, SAVP-VAE and \ac{SV2P}.
On the other hand, because video predictions made by \ac{CDNA} have changing dynamics, starting with good resolution and transitioning to blurry images as time advances, it is more difficult for the action inference model to learn to identify actions.

The performance of the action inference on predictions made by different models indicates, based on figures \ref{fig:r2_plots} and \ref{fig:l1_plots} and on \cref{table:r2_l1_fvd}, that the model that is better encoding action features and would therefore be the most suited in robotic planning problems is \ac{SV2P}, closely followed by SAVP-VAE, implying that conditioning on stochastic variables is beneficial but the introduction of the adversarial loss for better image quality removes attention from optimal encoding of action features. These models even outperform the ground truth oracle, supporting the author's argument that the stochastic variables should be accounting for non observable aspects of the scene and that some blurring of the background may actually help the inference network focus on the action features.
On the other hand, the action-free \ac{SVG-LP} model has an $R^2$ score of approximately 0 and an \ac{MAE} score of $0.163$ which corresponds to the variance of the data.
This indicates, as observed in section \ref{sec:experiments_qualitative}, that the inference model is unable to identify the actions and limits itself to predicting a constant average.
The origin of this result is that an action-free stochastic model from which a single prediction is sampled, may produce a future that is different from  the  ground truth, causing recognized actions to not match the targets and preventing the model from learning a meaningful mapping during training.

In general, and as reported by~\cite{unterthiner2018towards}, \ac{PSNR} and \ac{SSIM} present a very high correlation as both of them are based on frame by frame comparisons with the original data.
Furthermore, as most \ac{VP} models use an $\ell_2$ term in the loss function, these are biased metrics. We also verify that despite being the best ranked model under our metric, \ac{SV2P} ranks $3^{rd}$ in \ac{FVD}, with a score close to that of \ac{SVG-LP} which for being action-free has the worst $R^2$ score. These results show that despite there being some correlation between our metric and previously proposed ones, it is not as evident as correlation among existing metrics.

\begin{table}[h]
 \caption{FVD, R2 and MAE scores for each video predic- tion model (lower is better for FVD).}
  \centering
  \begin{tabular}{lcccc}
    \toprule
    \cmidrule(r){1-2}
    Model & FVD & $R^2$ & MAE \\
    \midrule
    CDNA     & 943.5 & 0.4339  & 0.0111 \\
    SAVP     & 738.3 & 0.5953  & 0.0092 \\
    SAVP-VAE & \textbf{409.8}  & 0.8427  & 0.0056 \\
    SV2P     & 691.1 & \textbf{0.8719}  & \textbf{0.0049} \\
    SVG      & 728.2 & -0.0582 & 0.0160 \\
    Oracle   & 0.0 & 0.8203 & 0.0058\\
    \bottomrule
  \end{tabular}
  \label{table:r2_l1_fvd}
\end{table}

\subsection{Qualitative Comparison}
\label{sec:experiments_qualitative}

\begin{figure}[h]
 \includegraphics[scale=0.23]{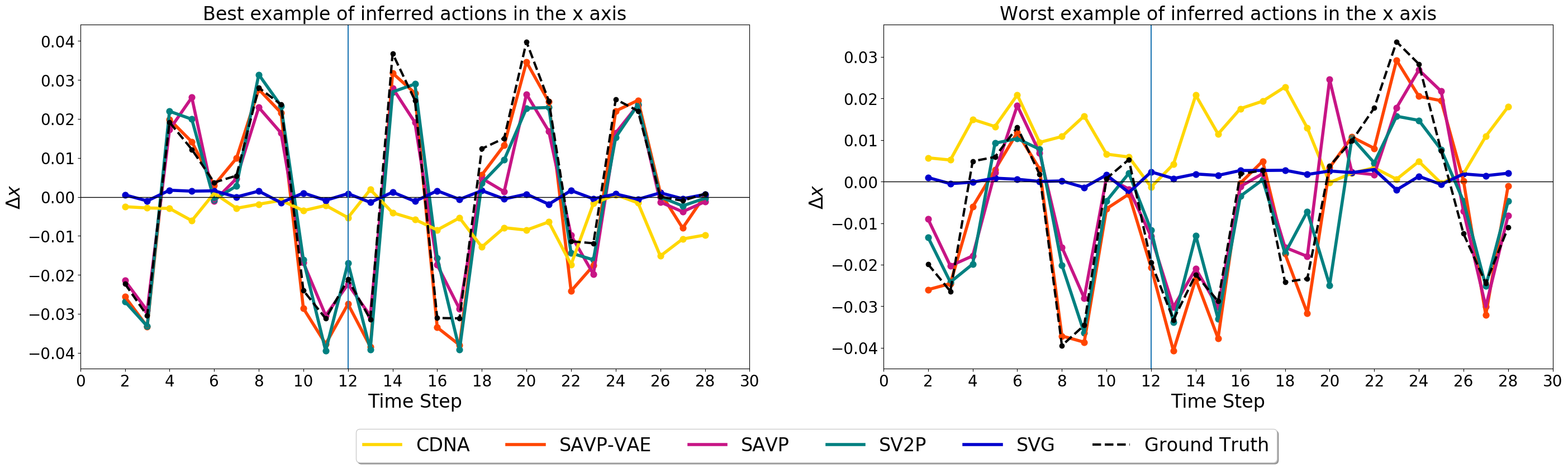}
 \centering
 \caption{Best and worst examples of inferred $\Delta x$.}
 \label{fig:x_reg}
\end{figure}

To better understand how the $R^2$ and MAE results reflect on action inference in practice, examples of actual inferred action sequences are showed against the ground truth in figure \ref{fig:x_reg}.
We find the best and worst $R^2$ score on action inference from video predictions from any \ac{VP} model and then plot the inferred actions from that video sequence.
The best performing model in the selected examples was SAVP-VAE, while the worst was SVG. A visual interpretation of the extent to which the inference network is successful at recognizing the actions is in conformance with the results of \cref{sec:experiments_quantitative}, with the SAVP-VAE and \ac{SV2P} models performing the best, followed by SAVP and CDNA.

\section{Conclusions and Future Work}
\label{sec:conclusions}

In this work we proposed a novel method for evaluating the quality of video prediction models from a robotic standpoint. We compared different existing video prediction models using our metric, showing that good performance on metrics that mirror human perception of quality does not necessarily imply that the model holds a good representation of action-effect. 
In future we plan to introduce better datasets that include states of the environment other than gripper position, such as objects positions and speeds, allowing the assessment of models based more comprehensive states. Developing action or object-state aware cost functions for training video prediction models is another possible future line of research.

\section*{Acknowledgements}

This work is partially supported by the Portuguese Foundation for Science and Technology (FCT) project [UID/EEA/50009/2019].


{\small
\bibliographystyle{unsrt}
\bibliography{references}
}

\end{document}